\documentclass[letterpaper]{article} % DO NOT CHANGE THIS
\usepackage{aaai24}  % DO NOT CHANGE THIS
\usepackage{times}  % DO NOT CHANGE THIS
\usepackage{helvet}  % DO NOT CHANGE THIS
\usepackage{courier}  % DO NOT CHANGE THIS
\usepackage[hyphens]{url}  % DO NOT CHANGE THIS
\usepackage{graphicx} % DO NOT CHANGE THIS
\usepackage{natbib}  % DO NOT CHANGE THIS AND DO NOT ADD ANY OPTIONS TO IT
\usepackage{caption} % DO NOT CHANGE THIS AND DO NOT ADD ANY OPTIONS TO IT
\usepackage{algorithm}
\usepackage{algorithmic}

\usepackage{multirow}
\usepackage{multicol} 
\usepackage{amsmath} % for \mathbb
\usepackage{amssymb} % for Bolded symbol in equation
\usepackage{booktabs} % for toprule bottomrule
\usepackage{xcolor}
\usepackage{adjustbox}

\usepackage{newfloat}
\usepackage{listings}
\DeclareCaptionStyle{ruled}{labelfont=normalfont,labelsep=colon,strut=off} % DO NOT CHANGE THIS
\floatstyle{ruled}
\newfloat{listing}{tb}{lst}{}
\floatname{listing}{Listing}
\title{FM-OV3D: Foundation Model-based Cross-modal Knowledge Blending for Open-Vocabulary 3D Detection}
\author{
    Dongmei Zhang\textsuperscript{\rm 1}\equalcontrib,
    Chang Li\textsuperscript{\rm 1}\equalcontrib,
    Ray Zhang\textsuperscript{\rm 2}\equalcontrib,
    Shenghao Xie\textsuperscript{\rm 3}\\
    Wei Xue\textsuperscript{\rm 4},
    Xiaodong Xie\textsuperscript{\rm 1},
    Shanghang Zhang\textsuperscript{\rm 1}\thanks{Corresponding authors.}
}
\affiliations{
    \textsuperscript{\rm 1}National Key Laboratory for Multimedia Information Processing, School of Computer Science, Peking University\\
    \textsuperscript{\rm 2}Shanghai AI Lab,
    \textsuperscript{\rm 3}Wuhan University,
    \textsuperscript{\rm 4}Hong Kong University of Science and Technology \\
    dmzhang@stu.pku.edu.cn, \{1900012977, donxie, shanghang\}@pku.edu.cn,\\  
    xieshenghao@whu.edu.cn, xuewei.x@gmail.com
}

\usepackage{bibentry}
\begin{document}

\maketitle

\begin{abstract}

The superior performances of pre-trained foundation models in various visual tasks underscore their potential to enhance the 2D models' open-vocabulary ability. Existing methods explore analogous applications in the 3D space. However, most of them only center around knowledge extraction from singular foundation models, which limits the open-vocabulary ability of 3D models. We hypothesize that leveraging complementary pre-trained knowledge from various foundation models can improve knowledge transfer from 2D pre-trained visual language models to the 3D space.
In this work, we propose \textbf{FM-OV3D}, a method of \textbf{F}oundation \textbf{M}odel-based Cross-modal Knowledge Blending for \textbf{O}pen-\textbf{V}ocabulary \textbf{3}D \textbf{D}etection, which improves the open-vocabulary localization and recognition abilities of 3D model by blending knowledge from multiple pre-trained foundation models, achieving true open-vocabulary without facing constraints from original 3D datasets. 
Specifically, to learn the open-vocabulary 3D localization ability, we adopt the open-vocabulary localization knowledge of the Grounded-Segment-Anything model.
For open-vocabulary 3D recognition ability, We leverage the knowledge of generative foundation models, including GPT-3 and Stable Diffusion models, and cross-modal discriminative models like CLIP.
The experimental results on two popular benchmarks for open-vocabulary 3D object detection show that our model efficiently learns knowledge from multiple foundation models to enhance the open-vocabulary ability of the 3D model and successfully achieves state-of-the-art performance in open-vocabulary 3D object detection tasks. Code is released at~\url{https://github.com/dmzhang0425/FM-OV3D.git}.

\end{abstract}

\section{Introduction}
\label{introduction}
Open-vocabulary ability refers to the ability of models to generate or understand concepts that have not been explicitly included in training datasets. 
Pre-trained foundation models' high performances in various 2D open-vocabulary visual tasks demonstrate their strong open-vocabulary abilities~\cite{liang2023open, liu2023grounding}.
However, utilizing vision-text pairs in training to enable such ability in 3D models is challenging, since it is difficult to collect a sizable dataset of 3D point clouds paired with texts.

The knowledge embedded in pre-trained foundation models can potentially enhance 3D models. Despite the differences in modalities between 3D point-cloud and 2D images, they both share visual information about objects. There have been efforts to investigate how to transfer knowledge from 2D to 3D models, leading to various distinct methods~\cite{zhang2022pointclip,zhu2023pointclip,zhang2023learning}. 

However, many of them primarily extract knowledge from individual models. Considering the disparities in training objectives, model architectures, and training data among various models, pre-trained models' knowledge, abilities, or perception of the world may exhibit diversity. This diversity can potentially enhance the open-vocabulary ability of 3D models complementarily. For example, different from the contrastive vision-language knowledge in CLIP~\cite{CLIP}, SAM~\cite{SAM} is designed to segment all objects in an image, providing information about their positions and sizes. Moreover, when vocabulary is involved in training open-vocabulary models to correlate visual and textual features, existing methods only use predefined class lists or captions, failing to provide rich information about the classes themselves, thereby limiting recognition performance. As a text-generative model, GPT-3~\cite{GPT} has a rich understanding of various classes and can serve as a source of textual knowledge. Therefore, we hypothesize that harnessing complementary pre-trained knowledge from different models can facilitate knowledge transfer from 2D pre-trained models to 3D space.

In this work, we propose FM-OV3D, a method of \textbf{F}oundation \textbf{M}odel-based Cross-modal Knowledge Blending for \textbf{O}pen-\textbf{V}ocabulary \textbf{3}D \textbf{D}etection, which improves the open-vocabulary localization and recognition ability of 3D models by incorporating knowledge from diverse pre-trained foundation models, without requiring any manual annotations.
Specifically, to train the open-vocabulary localization ability of 3D models, we utilize the object localization knowledge within the Grounded-Segment-Anything model to generate 2D bounding boxes.
To enhance the open-vocabulary recognition ability of 3D models, we associate the semantics among three different modalities of the same class: point cloud features from the 3D detector, CLIP extracted textual features of GPT-3 generated language prompts, and CLIP extracted visual features of Stable Diffusion-generated 2D visual prompts and point clouds' paired images. 
We perform open-vocabulary object detection in testing by comparing point cloud features and text features in a common feature space.
Moreover, our method can be applied to any manually selected open-vocabulary training set since our GPT-3 language prompts and Stable Diffusion visual prompts can be generated regarding any selected class. The major contributions of our work include:
\begin{itemize}
    \item We propose that leveraging complementary pre-trained knowledge from various foundation models can facilitate knowledge transfer from 2D pre-trained visual language models to the 3D space.
    \item We propose FM-OV3D, a method of foundation model-based cross-modal knowledge blending for open-vocabulary 3D detection, which incorporates knowledge of various foundation models to enhance the open-vocabulary localization and recognition ability of 3D models without requiring any manual annotation, which can be easily transferred to any 3D dataset.
    \item Experiments conducted on two public and commonly used open-vocabulary 3D point-cloud object detection datasets achieve \textit{state-of-the-art} performances, demonstrating that our method is effective.
\end{itemize}

\section{Related Work}
\label{related}

\subsection{Pre-trained Foundation Models}
Pre-trained foundation models are trained on massive amounts of data on a pre-defined proxy task. Models learn statistical structures and grasp the intrinsic links within training data, acquiring extensive knowledge. Large language models (LLMs) like GPT-3~\cite{GPT} are trained on a vast collection of internet text in self-supervised learning. They can generate human-like language responses and have been applied to various natural language processing downstream tasks~\cite{NER, textStr}. SAM~\cite{SAM} and PerSAM~\cite{zhang2023personalize} successfully incorporate visual content-relevant knowledge and have demonstrated high performances on various tasks~\cite{samformed1, samformed2}. However, these models' scope of knowledge is limited by insufficient training data across various modalities, training methods, and proxy task types used in training. As a result, existing pre-trained large models' applicability to downstream tasks is limited.

Recent research explores combining these pre-trained foundation models in various modalities. For example, CaFo~\cite{zhang2023prompt} cascades a variety of pre-trained foundation models to achieve better image classification performance. Grounding DINO~\cite{liu2023grounding}, which blends the knowledge of DINO with textual prompts, has state-of-the-art results in zero-shot settings.
Grounded-SAM~\cite{SAM,liu2023grounding}, which combines Grounding DINO~\cite{liu2023grounding} with SAM~\cite{SAM}, enhances detection and segmentation abilities simultaneously. However, constrained by fused models' limited modalities, there are still multi-modal problems, for instance, object detection and segmentation in 3D scenarios, yet to be explored.

\subsection{Open-Vocabulary 2D/3D Object Detection}
Open-vocabulary detection requires models to localize and recognize novel classes with training on only base classes~\cite{de2022structural, bangalath2022bridging,rahman2020improved, rahman2020zero, OpenVocabularyDef}.
Typically, knowledge of novel classes is indirectly implicated by cues from other modalities, for example, textual cues.
To enhance open-vocabulary detection capabilities, some studies explore rich image-text pairs' semantics' extraction~\cite{OpenVocabularyDef}.
Some works~\cite{rahman2020improved, rahman2020zero, OpenVocabularyDef} replace visual detectors' classification layer with a visual-textual embedding to achieve robust performance in open-vocabulary settings.

In 3D point cloud detection tasks, directly applying visual-language pre-trained models faces the challenge of acquiring large-scale point cloud-text pairs and the gap between image and point cloud modalities. Existing works seek solutions in utilizing foundation models' knowledge on 3D tasks. PointCLIP series~\cite{zhang2022pointclip,zhu2023pointclip,guo2022calip} use CLIP to process multi-view images projected from 3D modality, and Point-Bind\&Point-LLM~\cite{guo2023point} leverage LLMs and multi-modality semantics to achieve zero-shot 3D analysis. However, CLIP and LLMs are not trained for localization, and the localization in 2D space and 3D space differs significantly, which limits the localization capability of this method. Lu \textit{et al.}~\cite{lu2022open} expands the 3D detector's vocabulary with ImageNet1K~\cite{ImageNet}. 
Concurrently, Lu \textit{et al.}~\cite{lu2023open} proposes a divide-and-conquer strategy to connect textual information with point clouds.
They might limit the 3D detector's generalization ability originating from the dataset applied.
In this paper, we leverage multiple pre-trained models' knowledge from textual and image modalities, requiring no human-annotated data, enhancing our model's detection performance in open-vocabulary settings.

\section{Methodology}
\label{method}

\begin{figure*}[t]
  \centering
  \includegraphics[width=\linewidth]{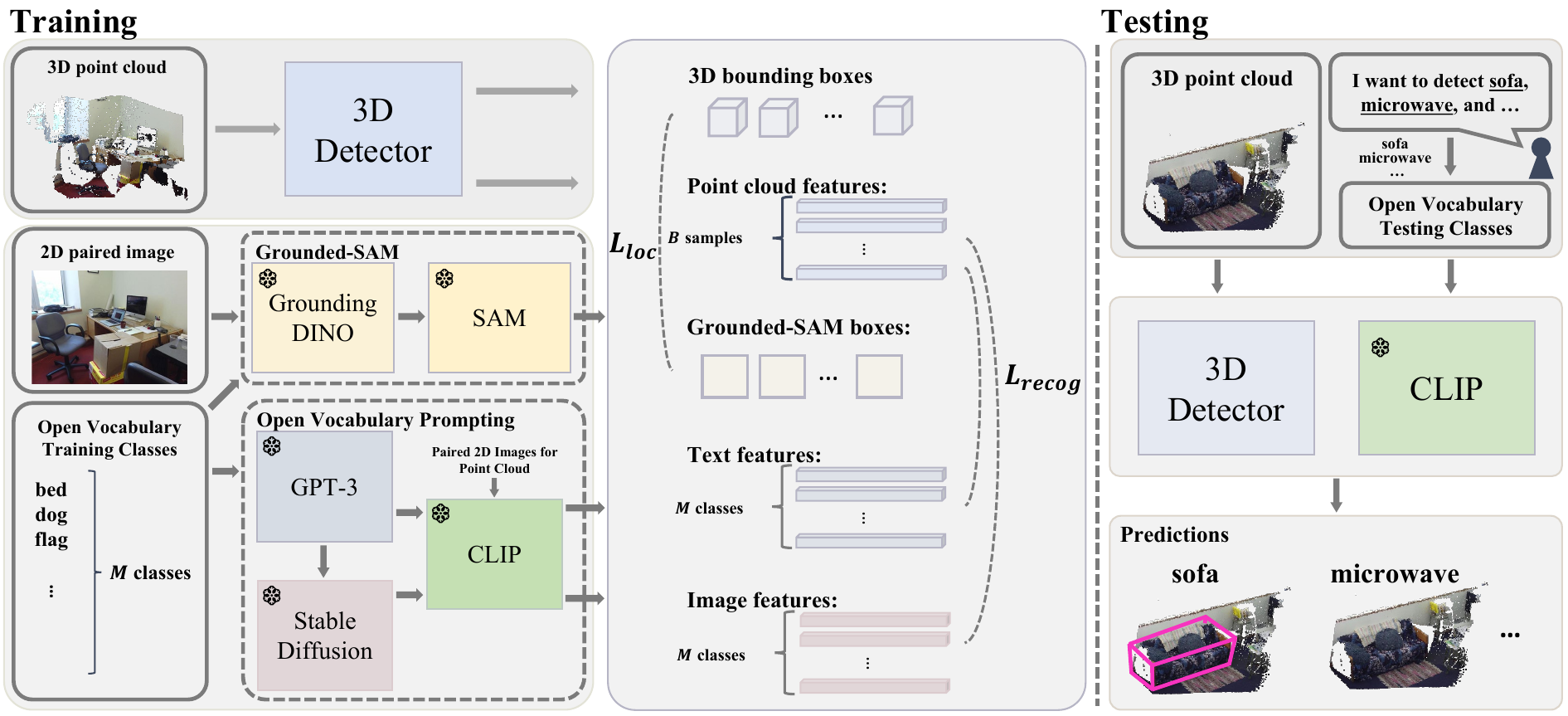}
  \caption{\textbf{The pipeline of FM-OV3D.} Given raw point clouds, corresponding 2D images, and training vocabulary, we train our model leveraging extensive knowledge from pre-trained foundation models, without requiring any annotations. $M$ and $B$ represent the size of the training vocabulary and the number of point cloud sample in a training batch, respectively. $L_{loc}$ aims at improving our 3D detector’s localization ability, while $L_{recog}$ is designed to improve its recognition ability. Point cloud visualization and its paired 2D image are selected from the SUN RGB-D~\cite{SUNRGBD} dataset, while training vocabulary is from the LVIS~\cite{lvis} dataset. `*' means the model is frozen.}
  \label{fig:method}
\end{figure*}

As shown in Figure~\ref{fig:method}, during training, we take raw point clouds, corresponding 2D images, and training vocabularies as input. During testing, the requirement for 2D images is eliminated, and the model relies only on testing vocabulary and the raw 3D point clouds.

Our 3D detector is required to predict the 3D bounding boxes transformed from the 2D results of Grounded-SAM to improve the open-vocabulary localization ability.
Regarding 3D open-vocabulary recognition ability, we blend the knowledge of GPT-3, Stable Diffusion, and CLIP. 
We conduct our recognition loss utilizing 3D features extracted by 3D detector, 2D features, and textual features extracted by CLIP, leveraging CLIP's rich cross-modal knowledge. Details are explained in the following sections.

\subsection{Open-Vocabulary 3D Localization}
\label{section:localization}

For the 3D detector's open-vocabulary localization ability, we employ the Grounded-Segment-Anything model, denoted as Grounded-SAM, to generate 3D bounding boxes for each point-cloud data and require our 3D detector to predict them. 

After being prompted with a training vocabulary, Grounded-SAM generates 2D detection boxes on 2D images corresponding to a set of images for a given point cloud. The training vocabulary consists of the LVIS~\cite{lvis} dataset that encompasses classes represented in text. From this dataset, a subset of $M$ classes $\{a_{1},\ldots, a_{M}$ \} is selected for prompting. 
It's noteworthy that we are agnostic about the classes of the 2D detection boxes.
The process of generating 2D bounding boxes for image $I$ is given by:
\begin{equation}
Box2D_{I} = GroundedSAM(a_{1},\ldots,a_{M}, I)
\label{eq1}
\end{equation}
where $Box2D_{I} \in \mathbf{R}^{4}$ represents the 2D boxes generated by Grounded-SAM. 
We then project these 2D bounding boxes via projection $K$ to 3D space and perform clustering to tighten the 3D bounding boxes.
\begin{equation}
Box3D_{I} = Cluster(Box2D_{I} \circ K^{-1})
\label{eq3}
\end{equation}
where $K$ is the projection matrix, which is provided in the datasets, and $Box3D_{I} \in \mathbf{R}^{7}$ represents the transformed 3D bounding boxes. $Clustering$ is a density-based clustering approach performed on points inside the projected bounding box to eliminate irrelevant outliers. 
% The full details of clustering can be accessed in the appendix. 

We supervise our 3D detector's predicted 3D bounding boxes $Box3D_{pc} \in  \mathbf{R}^{7}$ by above matched 3D bounding boxes $Box3D_{I}$. We compute bounding box regression loss following 3DETR~\cite{3DETR} demonstrated by Equation~\ref{eq4}. 
\begin{equation}
L_{loc} = \sum_{I=1}^{P} L_{box}^{3D}(Box3D_{I}, Box3D_{pc})
\label{eq4}
\end{equation}
where $L_{box}^{3D}$ represents regression loss between $Box3D_{pc}$ and the target $Box3D_{I}$. $P$ represents the number of matched bounding boxes in training.
By minimizing the value of $L_{loc}$, we enhance our 3D detector's open-vocabulary localization ability without requiring any annotation.

\subsection{Open-Vocabulary 3D Recognition}
We improve our 3D detector's open-vocabulary recognition ability by blending knowledge of single-modal generative foundation models, including GPT-3 and Stable Diffusion models, and cross-modal discriminative models like CLIP. Specifically, we utilize GPT-3 to generate rich textual prompts and the Stable Diffusion model to generate rich 2D visual prompts, then use CLIP to extract their features. 
 
Then we blend their knowledge by aligning object class semantics across three modalities: point cloud, images, and texts.

\paragraph{Text Prompt Generation}
GPT-3, with 175 billion parameters, is trained on a substantial amount of internet text in a self-supervised manner. 
We utilize its ability to generate rich, detailed, human-like language descriptions on training vocabulary.

For every training class, we prompt GPT-3 to generate detailed descriptions. We adopt existing templates from~\cite{CuPL} and prompt GPT-3 with ten rounds each, including \textit{“Describe what \{class\} look like”}, \textit{“How can you identify \{class\} ?”}, \textit{“What does \{class\} look like?”}, \textit{“Describe an image from the internet of \{class\} ”} and \textit{“A caption of an image of \{class\}:”}. 
We denote generated text prompts of $M$ training classes as $\{T_{1},\ldots, T_{M}\}$, and the overall text prompts as $T$. The prompts generated by GPT-3 contain extensive interpretations of semantic concepts, thus providing high-quality, diverse knowledge in textual modality.
\paragraph{2D Visual Prompt Generation}
We generate rich 2D images to provide our 3D model with abundant visual representations of open-vocabulary classes, broadening the vocabulary in the original 3D dataset. 
Stable Diffusion has an extensive textual-visual understanding and can generate synthesized 2D images according to language prompts.
Therefore, utilizing GPT-generated detailed descriptions $T_{1},\ldots, T_{M}$ for training vocabulary, we generate corresponding 2D images $SI$. 
\begin{equation}
    SI_{i} = SD(T_{i}), \ \
    SI = SD(T)
\label{eq6}
\end{equation}
where $i$ ranges from $1$ to $M$, $SI$ denotes all the 2D visual prompts generated for $M$ classes in training vocabulary. 
Given any training vocabulary, we can expand the training data in 2D vision and language modalities utilizing our method without requiring any human annotations, tackling the problem of limited represented classes in annotated 3D datasets.

\paragraph{Knowledge Blending}
The knowledge of GPT-3, Stable Diffusion, and CLIP are blended by aligning object class semantics across three modalities: point cloud features from our 3D detector, CLIP-extracted 2D image features of Stable-Diffusion generated 2D visual prompts and CLIP-extracted text features of textual prompts.

After blending, our 3D detector is trained to grasp the intrinsic links between visual objects in 3D modality and semantic concepts in 1D text modality.
We first project our 3D detector's predicted bounding boxes $Box3D_{pc}$ onto paired 2D image via dataset-provided projection matrix $K$, then get image crops $I_{pc}$.
The point cloud ROI features within $Box3D_{pc}$ are also extracted, denoted as $F_{pc}$.
Exploiting CLIP's visual-textual knowledge, we use CLIP to extract GPT-generated prompts $T$'s textual feature $F_{t}$, Stable Diffusion-generated 2D image $SI$'s features $F_{2D_{SI}}$, and image crops $I_{pc}$'s features $F_{2D_{pc}}$. The combination of $F_{2D_{SI}}$ and $F_{2D_{pc}}$ is represented as $F_{2D}$. The recognition loss among point cloud features, text features, and image features is given by:
\begin{equation}
L_{recog} = L_{cl}(F_{pc}, F_{2D}) + L_{cl}(F_{pc}, F_{t})
\label{eq7}
\end{equation}

Specifically, given a batch of features with size $B$ of 3D point cloud samples, $L_{cl}$ following~\cite{ContrastiveLoss} can be computed as follows:
\begin{equation}
L_{cl}(F_{1}, F_{2}) =  -\frac{1}{B}\sum_{b=1}^{B} \log(\frac{f(b, positive)}{f(b)})
\label{eq8}
\end{equation}
$F_{1}$ can be $F_{pc}$ and $F_{2}$ can be $F_{2D}$ or $F_{t}$ in Equation~\ref{eq7}. $f(b, positive)$ and $f(b)$ for every sample $f_{b}$ in the batch can be computed as follows:
\begin{equation}
\begin{aligned}
   f(b, positive) = \sum_{j=1}^{n}exp(f_{b}^{'}f_{j}/\tau),\\
    f(b) = \sum_{k=1}^{B}exp(f_{b}^{'}f_{k}/\tau) 
\end{aligned}
\label{eq9}
\end{equation}
where $\tau$ is the temperature parameter, $n$ is the number of positive samples.

The total loss function of our 3D detector can be computed as Equation~\ref{eq10}:
\begin{equation}
L = L_{loc} + L_{recog}
\label{eq10}
\end{equation}

\section{Experiments}
\label{experiments}

\begin{table*}[t]
\centering
\caption{Detection results ($AP_{25}$) on \textbf{SUN RGB-D} dataset. We report the accuracy of different classes and their mean score. * denotes our annotation-free version.}
\begin{center}
% \begin{adjustbox}{max width=5.5in}
\resizebox{\linewidth}{!}{
\begin{tabular}{ccccccccccccc}
\toprule\noalign{\smallskip}
Method & toilet & bed & chair & bathtub & sofa & dresser & scanner & fridge & lamp & desk & mean \\
\noalign{\smallskip}
\cmidrule(lr){1-12}
GroupFree3D~\cite{liu2021group} & 0.23 & 0.04 & 1.25 & 0.03 & 0.21 & 0.21 & 0.14 & 0.10 & 0.03 & 3.02 & 0.53 \\
VoteNet~\cite{qi2019deep} & 0.12 & 0.05 & 1.12 & 0.03 & 0.09 & 0.15 & 0.06 & 0.11 & 0.04 & 2.10 & 0.39 \\
H3DNet~\cite{zhang2020h3dnet} & 0.24 & 0.10 & 1.28 & 0.05 & 0.22 & 0.22 & 0.13 & 0.14 & 0.03 & 6.09 & 0.85 \\
3DETR~\cite{3DETR} & 1.57 & 0.23 & 0.77 & 0.24 & 0.04 & 0.61 & 0.32 & 0.36 & 0.01 & 8.92 & 1.31 \\
\cmidrule(lr){1-12}
OS-PointCLIP~\cite{zhang2022pointclip} & 7.90 & 2.84 & 3.28 & 0.14 & 1.18 & 0.39 & 0.14 & 0.98 & 0.31 & 5.46 & 2.26 \\
OS-Image2Point~\cite{xu2021image2point} & 2.14 & 0.09 & 3.25 & 0.01 & 0.15 & 0.55 & 0.04 & 0.27 & 0.02 & 5.48 & 1.20 \\
Detic-ModelNet~\cite{zhou2022detecting} & 3.56 & 1.25 & 2.98 & 0.02 & 1.02 & 0.42 & 0.03 & 0.63 & 0.12 & 5.13 & 1.52 \\
Detic-ImageNet~\cite{zhou2022detecting} & 0.01 & 0.02 & 0.19 & 0.00 & 0.00 & 1.19 & 0.23 & 0.19 & 0.00 & 7.23 & 0.91 \\
\cmidrule(lr){1-12}
OV-3DETIC~\cite{lu2022open} & 43.97 & 6.17 & 0.89 & \bf45.75 & 2.26 & \bf8.22 & 0.02 & 8.32 & 0.07 & \bf14.60 & 13.03 \\
\cmidrule(lr){1-12}
\textbf{FM-OV3D*} & 32.40 & 18.81 & \bf27.82 & 15.14 & \bf35.40 & 7.53 & \bf1.95 & \bf9.67 & 13.57 & 7.47 & 16.98 \\
\textbf{FM-OV3D} & \bf55.00 & \bf38.80 & 19.20 & 41.91 & 23.82 & 3.52 & 0.36 & 5.95 & \bf17.40 & 8.77 & \bf21.47 \\
\bottomrule
\end{tabular}}
% \end{adjustbox}
\end{center}
\label{table:sunrgbdAP}
\end{table*}

\begin{table*}[t]
\centering
\caption{Detection results ($AP_{25}$) on \textbf{ScanNet} dataset. We report the accuracy of different classes and their mean score. * denotes our annotation-free version.}
\begin{center}
% \begin{adjustbox}{max width=5.5in}
\resizebox{\linewidth}{!}{
\begin{tabular}{ccccccccccccc}
\toprule\noalign{\smallskip}
Method & toilet & bed & chair & sofa & dresser & table & cabinet & bookshelf & pillow & sink & mean \\
\noalign{\smallskip}
\cmidrule(lr){1-12}
GroupFree3D~\cite{liu2021group} & 0.63 & 0.52 & 1.25 & 0.52 & 0.20 & 0.59 & 0.52 & 0.25 & 0.01& 0.15 & 0.49 \\
VoteNet~\cite{qi2019deep} & 0.04 & 0.02 & 0.12 & 0.00 & 0.02 & 0.11 & 0.07 & 0.05 & 0.00 & 0.00 & 0.04 \\
H3DNet~\cite{zhang2020h3dnet} & 0.55 & 0.29 & 1.70 & 0.31 & 0.18 & 0.76 & 0.49 & 0.40 & 0.01 & 0.10 & 0.48 \\
3DETR~\cite{3DETR} & 2.60 & 0.81 & 0.90 & 1.27 & 0.36 & 1.37 & 0.99 & 2.25 & 0.00 & 0.59 & 1.11 \\
\cmidrule(lr){1-12}
OS-PointCLIP~\cite{zhang2022pointclip} & 6.55 & 2.29 & 6.31 & 3.88 & 0.66 & 7.17 & 0.68 & 2.05 & 0.55 & 0.79 & 3.09 \\
OS-Image2Point~\cite{xu2021image2point} & 0.24 & 0.77 & 0.96 & 1.39 & 0.24 & 2.82 & 0.95 & 0.91 & 0.00 & 0.08 & 0.84 \\
Detic-ModelNet~\cite{zhou2022detecting} & 4.25 & 0.98 & 4.56 & 1.20 & 0.21 & 3.21 & 0.56 & 1.25 & 0.00 & 0.65 & 1.69 \\
Detic-ImageNet~\cite{zhou2022detecting} & 0.04 & 0.01 & 0.16 & 0.01 & 0.52 & 1.79 & 0.54 & 0.28 & 0.04 & 0.70 & 0.41 \\
\cmidrule(lr){1-12}
OV-3DETIC~\cite{lu2022open} & 48.99 & 2.63 & 7.27 & 18.64 & \bf2.77 & \bf14.34 & \bf2.35 & 4.54 & 3.93 & 21.08 & 12.65 \\
\cmidrule(lr){1-12}
\textbf{FM-OV3D*} & 2.17 & 41.11 & \bf27.91 & \bf33.25 & 0.67 & 12.60 & 2.28 & \bf8.47 & 9.08 & 5.83 & 14.34 \\
\textbf{FM-OV3D} & \bf62.32 & \bf41.97 & 22.24 & 31.80 & 1.89 & 10.73 & 1.38 & 0.11 & \bf12.26 & \bf30.62 & \bf21.53 \\
\bottomrule
\end{tabular}}
% \end{adjustbox}
\end{center}
\label{table:scannetAP}
\end{table*}

In this section, we evaluate our FM-OV3D on widely used 3D detection datasets and analyze the incorporated foundation model's effects on open-vocabulary 3D detection models. We also discuss the influence of some key parameters.

\subsection{Datasets and Evaluation Metrics}

\paragraph{Datasets} We conduct experiments on public, widely used datasets \textbf{SUN RGB-D}~\cite{SUNRGBD} and \textbf{ScanNet}~\cite{Scannet} in 3D object detection tasks. The provided point-cloud data and corresponding images, together with the matrix $K$, are used in our method. 

\paragraph{Evaluation Metrics} We use mean Average Precision (AP), and Average Recall (AR) at IoU thresholds of 0.25 and 0.5, denoted as $mAP_{25}$, $mAP_{50}$, $AR_{25}$, and $AR_{50}$, as our primary metrics.

\subsection{Implementation Details}
We adopt LVIS~\cite{lvis} as our training vocabulary.
600 random classes sampled from our training vocabulary are used to prompt Grounded-SAM to generate 2D bounding boxes.
We adopt five templates as the commands for generating GPT-3 textual prompts and later compute the mean textual feature of each class, following~\cite{CuPL}. 
We apply the stable-diffusion-v1-4 model commanded by GPT-3 generated prompts. 
CLIP version ViT-B/32 is used for extracting features. We conduct our ablation studies on the SUN RGB-D dataset.

We train our model in 400 epochs. The base learning rate is set to 7e-4. We load 8 3D scenes onto each GPU in every batch.
We adopt 3DETR~\cite{3DETR} as the 3D detector. 
Experiments are conducted on two NVIDIA GeForce RTX 3090 GPUs and A100 SXM4 80GB GPUs.
In evaluation, we take our 3D detector's predicted 3D boxes as the localization result and the CLIP-predicted label of its corresponding 2D image crop as its label output.

\begin{table*}
\centering
\begin{center}
\caption{\textbf{Ablation study (\%) of pre-trained foundation models.} `GS' represents Grounded-SAM, and `SD' is short for Stable Diffusion. `Annotation' indicates whether 2D and 3D annotations are used.}
\label{table:multiAnnotation}
\begin{tabular}{cccccc}
\toprule\noalign{\smallskip}
~ & ~ & ~ & ~ & ~ & ~ \\
\noalign{\smallskip}
Models & Annotation & $mAP_{25}$ & $AR_{25}$ & $mAP_{50}$ & $AR_{50}$ \\
\noalign{\smallskip}
\cmidrule(lr){1-6}
\noalign{\smallskip}
GPT-3 & \checkmark & 18.09 & \bf53.87 & 1.88 & \bf11.58 \\
SD & \checkmark & 16.34 & 47.69 & 1.10 & 8.60 \\
GPT-3 + SD & \checkmark & \bf18.19 & 49.90 & \bf1.93  & 10.05 \\
\cmidrule(lr){1-6}
\noalign{\smallskip}
% GS & $\times$ & 16.47 & 55.59 & 1.84 & 11.47 \\
% GS + GPT-3 & $\times$ & 15.78 & 54.23 & 1.99 & 12.99 \\
% % GS + SD & $\times$ & 12.40 & 47.46 & 1.21 & 8.85 \\ 
% GS + GPT-3 + SD & $\times$ & 16.98 & 57.22 & 1.86  & 12.16 \\
GS & - & 16.47 & 55.59 & 1.84 & 11.47 \\
GS + GPT-3 & - & 15.78 & 54.23 & \bf1.99 & \bf12.99 \\
% GS + SD & $\times$ & 12.40 & 47.46 & 1.21 & 8.85 \\ 
GS + GPT-3 + SD & - & \bf16.98 & \bf57.22 & 1.86  & 12.16 \\
\bottomrule
\end{tabular}
\end{center}
\end{table*}

% \begin{table*}[t]
% \centering
% \caption{\textbf{Ablation Study (\%) of Pre-trained Foundation Models.} We ablate different pre-trained foundation models used in our method.}
% \begin{center}
% \begin{tabular}{ccccccc}
% \toprule\noalign{\smallskip}
% \multicolumn{3}{c}{{Pre-trained Models}} & ~ & ~ & ~ & ~ \\
% \noalign{\smallskip}
% \cmidrule(lr){1-3}
% \noalign{\smallskip}
% Grounded SAM & GPT-3 & Stable-Diffusion & $mAP_{25}$ & $mAP_{50}$ & $AR_{25}$ & $AR_{50}$ \\
% \noalign{\smallskip}
% \cmidrule(lr){1-7}
% \noalign{\smallskip}
% \multicolumn{1}{c}{\checkmark} & ~ & ~ & 16.47 & 1.84 & 55.59 & 11.47 \\
% ~ & \multicolumn{1}{c}{\checkmark} & ~ & 18.09 & 1.88 & 53.87 & 11.58 \\
% ~ & ~ & \multicolumn{1}{c}{\checkmark} & 16.34 & 1.10 & 47.69 & 8.60 \\
% \multicolumn{1}{c}{\checkmark} & \multicolumn{1}{c}{\checkmark} &{} & 15.78 & \bf1.99  & 54.23 & \bf12.99 \\
% \multicolumn{1}{c}{\checkmark} &{} & \multicolumn{1}{c}{\checkmark} & 12.40 & 1.21  & 47.46 & 8.85 \\ 
% {} & \multicolumn{1}{c}{\checkmark} & \multicolumn{1}{c}{\checkmark} & \bf18.19 & 1.93  & 49.90  & 10.05 \\
% \cmidrule(lr){1-7}
% \multicolumn{1}{c}{\checkmark} & \multicolumn{1}{c}{\checkmark} & \multicolumn{1}{c}{\checkmark} & 16.98 & 1.86  & \bf57.22  & 12.16 \\
% \bottomrule
% \end{tabular}
% \end{center}
% \label{table:multi}
% \end{table*}

\subsection{Performance on Open-Vocabulary 3D Object Detection}
Since no prior studies have addressed open-vocabulary 3D point cloud detection problems by avoiding the need for human annotations, we compare our model's performance with a 3D detection model~\cite{lu2022open} that utilizes human annotations and is exposed to open-set knowledge from other modalities. 
We select our open-testing classes following~\cite{lu2022open} and adopt their models discussed in an open set setting for comparison. 
We denote our model trained in an annotation-free setting as FM-OV3D*, and FM-OV3D represents the model trained only utilizing knowledge blending, utilizing Detic~\cite{zhou2022detecting} for 2D bounding box predictions.
Results of our experiments on SUN RGB-D and ScanNet are shown in Table~\ref{table:sunrgbdAP} and Table~\ref{table:scannetAP}. 

Our annotation-free model surpasses existing open-set 3D point cloud detector benchmarks, reaching $16.98\%$ on SUN RGB-D and $14.34\%$ on ScanNet in the $mAP_{25}$, demonstrating our model's outstanding performance on detecting 3D objects outside the training vocabulary, indicating its strong open-vocabulary ability. 
Furthermore, compared to OV-3DETIC~\cite{lu2022open}, which achieves strong performance by leveraging the knowledge in 2D image datasets, our model blends the knowledge from both textual and 2D visual modalities. Utilizing pre-trained models' generative knowledge allows our 3D detector to grasp the intrinsic links among three modalities, without exploiting knowledge from other datasets. 
Therefore, our method has no constraints originating from our leveraged cross-modal knowledge. Also, our method does not require human annotation in training.
Our outstanding experiment results on open-vocabulary testing classes indicate that by incorporating general representations learned by various foundation models, we can bridge the gap between the limited classes in annotated 3D datasets and real-world applications, improving the 3D detector's open-vocabulary ability. Our method can be applied to any selected open-vocabulary training set without utilizing data other than raw 3D point cloud data.

Our model FM-OV3D that only leverages our knowledge blending stage demonstrates significantly enhanced performance, outperforming previous methods by $8.44\%$ on SUN RGB-D and $8.88\%$ on ScanNet in terms of $mAP_{25}$. Despite the notable improvement in overall $mAP_{25}$ performance, 
we observe certain classes where FM-OV3D exhibits less satisfactory results. 
For example, the performance on the `scanner' of FM-OV3D falls short compared to FM-OV3D*. This suggests the potential for further enhancement of FM-OV3D by enriching its visual information related to open-vocabulary concepts.

\subsection{Ablation Study}

\begin{table}
\centering 
\caption{\textbf{Ablation study (\%) of the number of selected text prompts of each class.} $TP$ represents the number of GPT-3 generated text prompts selected in an individual class.}
\begin{center}
\begin{tabular}{ccccc}
\toprule\noalign{\smallskip}
TP & $mAP_{25}$ & $mAP_{50}$ & $AR_{25}$ & $AR_{50}$ \\
\noalign{\smallskip}
\cmidrule(lr){1-5}
\noalign{\smallskip}
12 & 11.66 & 1.10 & 45.37 & 7.67 \\
25 & 11.70 & \bf1.38 & \bf48.41 & 8.26 \\
51 & \bf12.50 & 1.16 & 46.73 & \bf8.65 \\
\bottomrule
\end{tabular}
\end{center}
\label{table:text}
\end{table}

\paragraph{The effect of pre-trained foundation models.} We investigate the effects of each pre-trained foundation model in our 3D detector's training, as shown in Table~\ref{table:multiAnnotation}.
`GS' denotes using the 2D bounding boxes generated by the Grounded-SAM model, `GPT-3' represents that we utilize GPT-3 to generate text prompts according to the open-vocabulary training classes, and `SD' denotes the 2D visual prompts generation utilizing the Stable Diffusion model. We apply bounding boxes generated by Detic~\cite{zhou2022detecting} as our 2D detection baseline, which are marked as annotation-needed in Table~\ref{table:multiAnnotation}.
The experimental results of our method are demonstrated in the last row.

Compared to the annotation-need model which is trained using only 2D visual prompts generated by Stable Diffusion or text prompts by GPT-3, the same model which is trained using both 2D visual prompts and text prompts shows better performance in $mAP_{25}$ and $mAP_{50}$. 
Similarly, when training the annotation-free model with both text and 2D visual prompts, its performance surpasses that of using either prompt alone or having no prompts at all. Specifically, the model incorporating both prompts achieves the best performance on the more stringent $mAP_{25}$ and $AR_{25}$ metrics. Meanwhile, the model exclusively utilizing text prompts attains the best $mAP$ and $AR$ when the threshold is set to 50. These results demonstrate the effectiveness of diverse semantic information from text-generative and cross-modal generative models even in the presence of annotations. It is reasonable since the method that only enlarges textual information lacks bridging between visual representations and textual knowledge. Although combining rich 2D visual prompts can enrich the original 3D dataset's visual data, the model lacks direct visual-textual cross-modal understanding, and GPT-3 includes rich semantics information on target concepts and is in a more direct form of the model's object recognition predictions.

We investigate the effect of bounding boxes generated by open-vocabulary classes-prompted Grounded-SAM. In comparison to using Detic~\cite{zhou2022detecting} for generating 2D bounding boxes during training our detector's localization abilities, our method demonstrates outstanding performance in $AR_{25}$ and $AR_{50}$ without relying on any manual annotations, indicating its strong ability to localize objects in 3D modalities accurately.

Overall, compared with models that ditch one component in our design, the results of our method demonstrate that with the open-vocabulary 2D grounding abilities, language generative knowledge, and 2D visual generative knowledge combined, the 3D detector can fully understand the concepts' representation in three modalities, achieving superior results in open vocabulary 3D detection tasks.

\begin{table}
\centering
\caption{\textbf{Ablation study (\%) of number of classes of 2D visual prompts.} $VP$ represents the number of selected Stable-Diffusion generated 2D visual prompts.}
\begin{center}
\begin{tabular}{ccccc}
\toprule\noalign{\smallskip}
VP & $mAP_{25}$ & $mAP_{50}$ & $AR_{25}$ & $AR_{50}$ \\
\noalign{\smallskip}
\cmidrule(lr){1-5}
\noalign{\smallskip}
0 & 14.09 & 2.00 & 50.21 & 12.56 \\
3 & 12.84 & \bf2.24 & 51.18 & \bf13.11 \\
5 & \bf15.49 & 2.13 & \bf53.48 & 11.92 \\
7 & 12.93 & 2.20 & 47.98 & 10.20 \\
10 & 13.76 & 1.73 & 52.14 & 12.05 \\
\bottomrule
\end{tabular}
\end{center}
\label{table:image}
\end{table}

\begin{figure}[t]
    \centering
    \includegraphics[width=\linewidth]{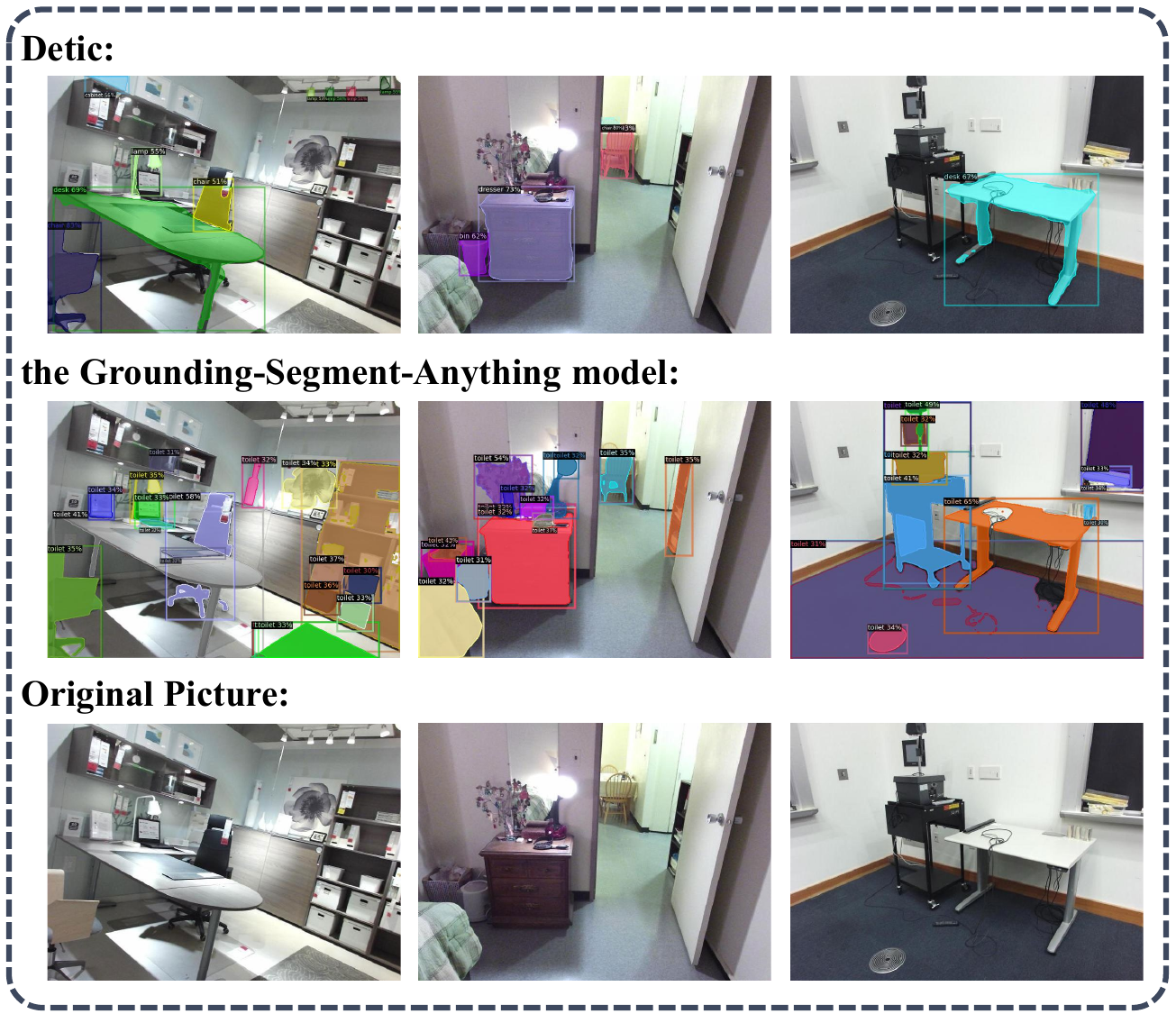}
    \caption{Bounding boxes generated by the Grounded-Segment-Anything model and Detic Original pictures are selected from the \textbf{SUN RGB-D} dataset.}
    \label{fig:SAMBoxVis}
\end{figure}

\begin{figure}[t]
  \centering
  \includegraphics[width=\linewidth]{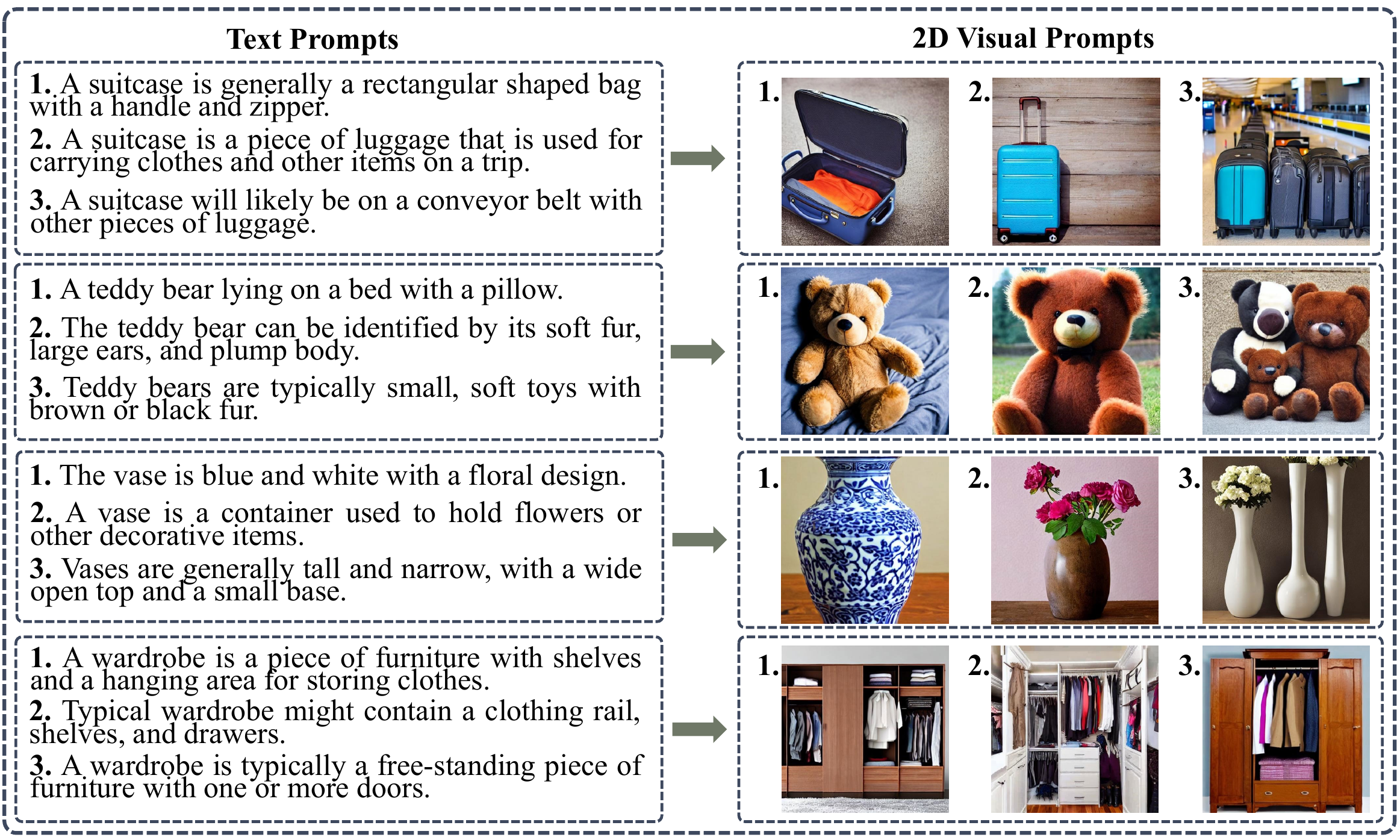}
  \caption{Visualization of GPT-generated prompts and their paired 2D visual prompts generated by Stable Diffusion on open vocabulary training dataset sampled from \textbf{LVIS}.}
  \label{fig:StableDiffusionVis}
\end{figure}

\paragraph{Number of selected text prompts of each class.} We generate multiple text prompts using GPT-3 in our method.
We study the influence of the number of generated text prompts for an individual class on the performance of our model. The results are shown in Table~\ref{table:text}. Considering training difficulty and time constraints, we evaluate the performance of 12, 25, and all text prompts. Experimental results indicate that utilizing all text prompts yields the highest $mAP_{25}$, demonstrating the effectiveness of rich textual features in enhancing the recognition capability of the detector. However, even when using only half of the text prompts (\textit{i.e.}, 25), the model still achieves the best $mAP_{50}$ and $AR_{25}$. Although reducing the number of text prompts has a certain impact on the model's recognition performance, it remains a competitive option as it achieves comparable performance with less generation and training burden. Further reduction in the number of prompts significantly affects the model's performance, underscoring the importance of incorporating text prompts.

\paragraph{Number of classes of used 2D visual prompts.} 
The performances of applying different numbers of classes of 2D visual prompts are shown in Table~\ref{table:image}. 
We report the best $mAP_{25}$ and $AR_{25}$ when applying 5 classes of visual prompts. Compared to using a smaller number of classes or ditching 2D visual prompts, the performance achieved with 5 classes demonstrates the effectiveness of 2D visual prompts in enhancing the model's recognition capability. The model's performance declines as the number of classes increases. This can be attributed to the introduction of excessive negative samples, which introduce noise and conflicts, hindering the model's ability to learn accurate and robust feature representations and increasing the difficulty of training. Therefore, incorporating 2D visual prompts is crucial for enhancing model performance, but the number of selected classes applied should be taken into consideration to achieve optimal results.

\subsection{Qualitative Analysis}

We compare the bounding boxes generated by Grounded-SAM to the 2D boxes generated by Detic~\cite{zhou2022detecting}. We assign the class prediction of boxes generated by Grounded-SAM to a random value since we do not further use these labels for predictions. In Figure~\ref{fig:SAMBoxVis}, objects localized by the Grounded-SAM are based on our prompted open-vocabulary classes, not facing constraints originating from the classes detected by pre-trained 2D detectors.
More objects are detected when Grounded-SAM is applied.
Since we use replaceable, open-vocabulary training classes to prompt the model, we achieve strong open-vocabulary ability.

We visualize GPT-3-generated text prompts and the paired 2D visual prompts generated by stable diffusion according to the GPT-3 text prompts. In Figure~\ref{fig:StableDiffusionVis}, we show that our visual prompts are diverse and variant in their direct representation of the commanded classes due to the rich GPT-3 generated descriptions, enriching the visual information and alleviating original 3D datasets' data insufficiency problem. 
Our visual prompts also successfully grasp the semantic concepts of the commanded classes, utilizing Stable Diffusion's vision-language knowledge.

\section{Conclusion}
\label{conclusion}
We demonstrate that leveraging complementary pre-trained knowledge from various foundation models can improve the knowledge transfer from 2D pre-trained foundation models to the 3D space, therefore enhancing the open-vocabulary ability of 3D models.
We propose FM-OV3D, a foundation model-based cross-modal knowledge blending for open-vocabulary 3D object detection method that correlates multi-modal knowledge from different foundation models onto the 3D modality without requiring any human annotation.
We train our 3D detector in aspects of localization and recognition. For open-vocabulary localization, we integrate the Grounded-Segment-Anything model's 2D knowledge by transforming its 2D bounding box predictions to supervise our model's localization results. 
For open-vocabulary recognition,
we blend the knowledge from pre-trained text and image generative models and cross-modal discriminative models with knowledge in 3D modality, bridging the gap between abundant real-world classes and the insufficiency of classes in 3D datasets.
We conduct experiments on SUN RGB-D and ScanNet datasets, and our experimental results demonstrate that our method is effective.

\section{Acknowledgments}
Shanghang Zhang is supported by the National Key Research and Development Project of China (No.2022ZD0117801). This work was supported by Emerging Engineering Interdisciplinary Youth Program at Peking University under Grant 7100604372.

\bibliography{aaai24}

\end{document}